\documentclass{article}
\usepackage{acra}
\usepackage{graphics}
\usepackage{graphicx}
\usepackage{subcaption}
\usepackage{booktabs} 
\usepackage{siunitx}
\usepackage{url}

\title{Virtual Traffic Lights for Multi-Robot Navigation: Decentralized Planning with Centralized Conflict Resolution}
\author{
    \textbf{Sagar Gupta$^1$, Thanh Vinh Nguyen$^1$, Thieu Long Phan$^1$, Vidul Attri$^1$, Archit Gupta$^1$,} \\
    \textbf{Niroshinie Fernando$^1$, Kevin Lee$^1$, Seng W. Loke$^1$, Ronny Kutadinata$^2$,} \\ 
    \textbf{Benjamin Champion$^1$ and Akansel Cosgun$^1$} \\
    $^1$Deakin University, Australia \\
    $^2$National Transport Research Organisation (NTRO), Australia \\
    guptasag@deakin.edu.au
}

\begin{document}

\maketitle

\begin{abstract}
We present a hybrid multi-robot coordination framework that combines decentralized path planning with centralized conflict resolution. In our approach, each robot autonomously plans its path and shares this information with a centralized node. The centralized system detects potential conflicts and allows only one of the conflicting robots to proceed at a time, instructing others to stop outside the conflicting area to avoid deadlocks. Unlike traditional centralized planning methods, our system does not dictate robot paths but instead provides stop commands, functioning as a virtual traffic light. In simulation experiments with multiple robots, our approach increased the success rate of robots reaching their goals while reducing deadlocks. Furthermore, we successfully validated the system in real-world experiments with two quadruped robots and separately with wheeled Duckiebots.
\end{abstract}

\section{Introduction}

Coordinating multiple robots to navigate shared spaces efficiently and without collisions has many applications in robotics, video games, and traffic control. Centralized and decentralized multi-agent coordination each have distinct trade-offs. Centralized systems can achieve optimal global solutions but suffer from computational complexity and a single point of failure~\cite{berndt2021centralized}. Decentralized systems offer greater scalability and robustness, but local decision-making can lead to suboptimal or conflicting behaviors, such as deadlocks~\cite{jingjin2016intractability}.

\begin{figure}
    \centering
    \includegraphics[width=0.99\linewidth]{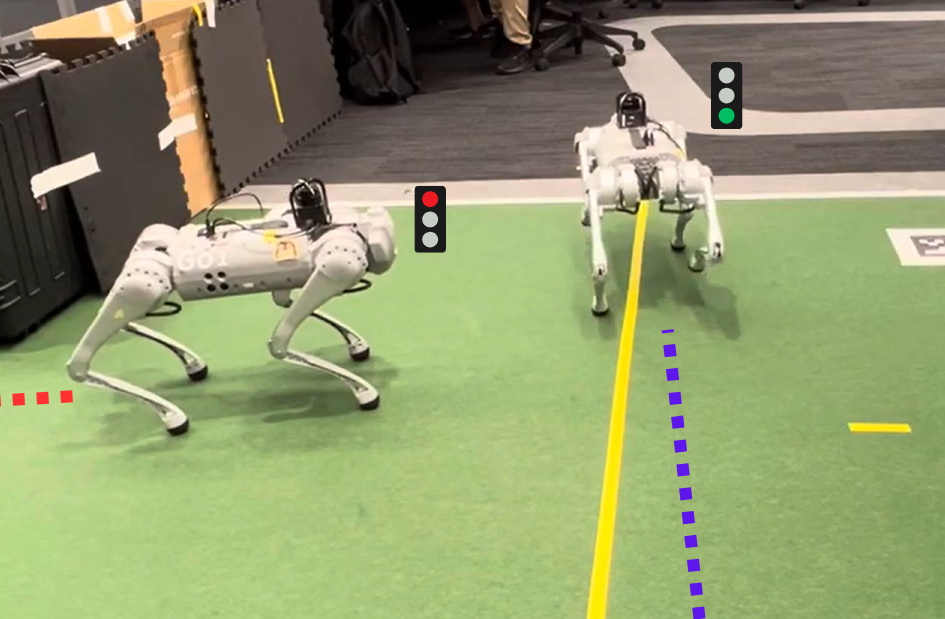}
    \caption{Our hybrid system combines decentralized planning with centralized conflict resolution. To prevent a deadlock, a central node issues a virtual red light to the quadruped on the left, allowing the other robot to pass safely through the conflict zone. Dotted lines represent each robot's traversed path.}
    \label{fig:mainimg}
\end{figure}

To address these limitations, we propose a hybrid coordination framework that combines decentralized motion planning with centralized conflict resolution. Each robot operates autonomously, independently planning its path, which reduces the central computational load as no single authority dictates detailed trajectories. The centralized system examines agent paths for conflicts; when it predicts one, it declares the area an intersection and issues a simple “Stop” command to conflicting robots, allowing only one to pass at a time. Figure~\ref{fig:mainimg} shows the kind of scenarios being addressed in this work.

The coordination system is agnostic of the robots' navigation stacks, as it operates on generic planned paths and produces commands to pause or resume navigation. This guidance facilitates cooperation among otherwise incompatible robots, making the framework suitable for real-world deployment. To demonstrate this, we have implemented the system in an identical manner across different navigation systems and communication frameworks.

Our contributions are three-fold:

\begin{itemize}
\item We propose a platform-agnostic, hybrid coordination framework that combines decentralized path planning with a lightweight, centralized conflict resolution mechanism functioning as a virtual traffic light.
\item We conduct a simulation study with 1,000 trials to quantitatively evaluate our system's performance against a purely decentralized baseline, measuring success rate, average speed, and path replans across varying numbers of robots.
\item We validate the framework's practical applicability and flexibility through real-world hardware demonstrations on two distinct platforms: dynamic quadruped robots and lane-following wheeled robots.
\end{itemize}

\section{Related Works}

Multi-Agent Path Finding (MAPF) problems attempt to solve conflicts between a group of robots, where each robot attempts to navigate from a unique starting point to a unique destination. These solutions are optimized for a cost function like time or energy~\cite{gunter2014history,atzmon}. MAPF is an NP-Hard problem~\cite{surynek2010optimization,jingjin2016intractability}, which is a computational problem that is at least as difficult as the hardest problems in the NP class, for which no known efficient algorithm exists. Prior works to solve the Multi-Agent Path Finding (MAPF) problem have been categorized into decentralized, centralized, hybrid, and learning-based coordination approaches.

Centralized approaches use a single supervisor which has global knowledge of the map and the states of the robots. This central node plans and coordinates the paths for the robots within the map such that each robot reaches its destination without colliding. ~\cite{effmulti} utilizes a central fleet manager that plans collision-free paths using Time Enhanced A* (TEA*).~\cite{intmulti} uses a combination of Genetic Algorithms (GA) and A*, which integrates fuzzy logic for obstacle avoidance and uses a cubic spline interpolation curve to reduce energy use.~\cite{supermulti} uses a centralized control law to navigate robots to their targets while maximizing efficiency of the path. Centralized multi-robot coordination is prevalent in industrial settings where task planning and scheduling is based on robot availability \cite{caloud1990indoor,berndt2021centralized}. 
\begin{figure}
    \centering
    \includegraphics[width=0.99\linewidth]{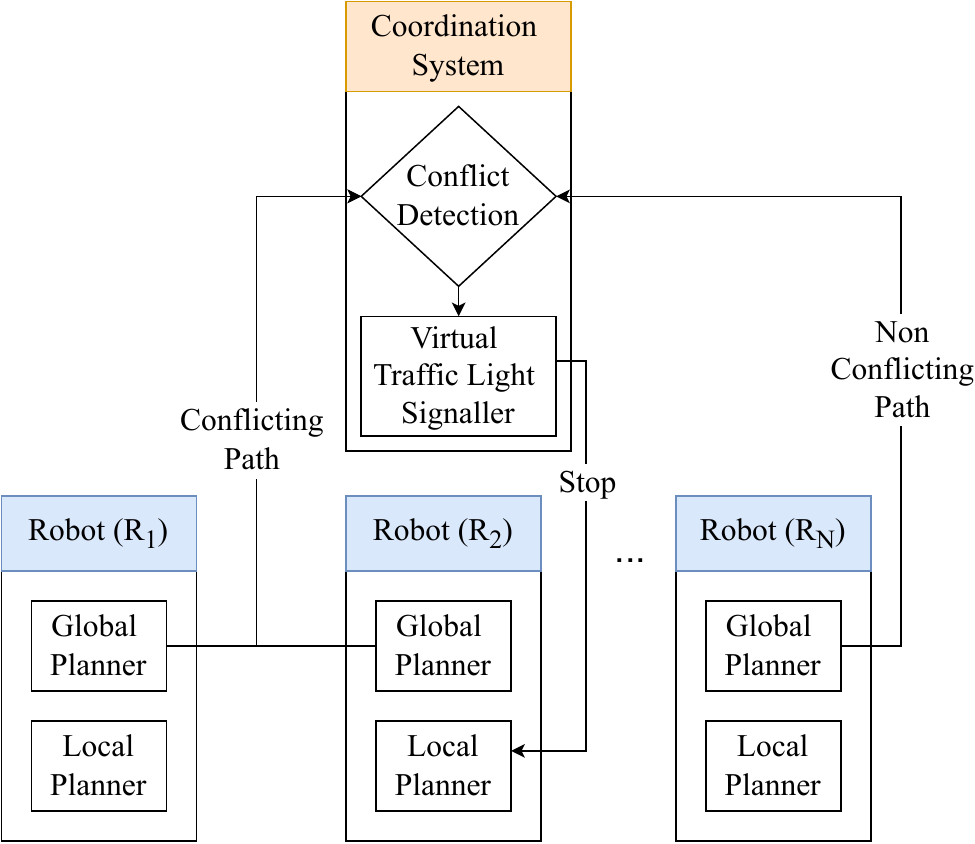}
    \caption{System overview with N robots, where only robots $R_{1}$ and $R_{2}$ have a conflicting path. The centralized coordination system allows $R_{1}$ to proceed and halts $R_{2}$ until $R_{1}$ has navigated through the conflicting intersection.}
    \label{fig:sysoverview}
\end{figure}
Decentralized coordination is used in large-scale systems with unreliable communication networks \cite{siefke2020robotic}. In this approach, each robot operates autonomously, making decisions based on information captured from sensors (implicit communication) or explicitly communicated information between robots \cite{flocchini2000distributed,iocchi2003distributed,jouandeau2012decentralized}. Deep Reinforcement Learning (DRL) can learn to coordinate robot fleets \cite{junyan2020voronoi,de2021decentralized}. Other learning-based strategies use a fuzzy inference system to refine paths generated by a global planner like D* Lite \cite{zagradjanin2021cloud}. A common characteristic of these methods is their reliance on prior system data, such as successful runs in simpler environments, to train the coordination policy \cite{kulathunga2021reinforcement}. 

\begin{figure*}[t]
    \centering
    \begin{subfigure}[b]{0.32\textwidth}
        \includegraphics[width=\textwidth]{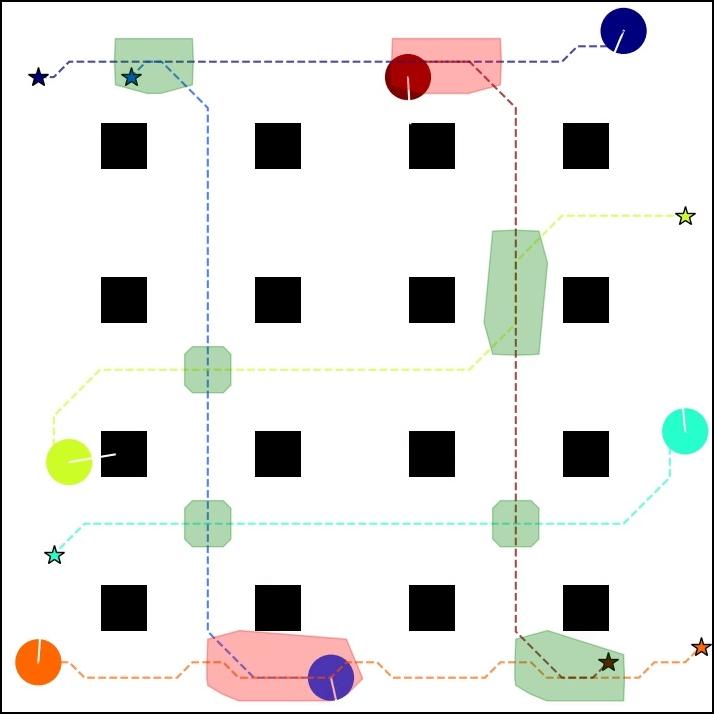}
        \caption{t=0s}
        \label{fig:t0}
    \end{subfigure}
    \hfill
    \begin{subfigure}[b]{0.32\textwidth}
        \includegraphics[width=\textwidth]{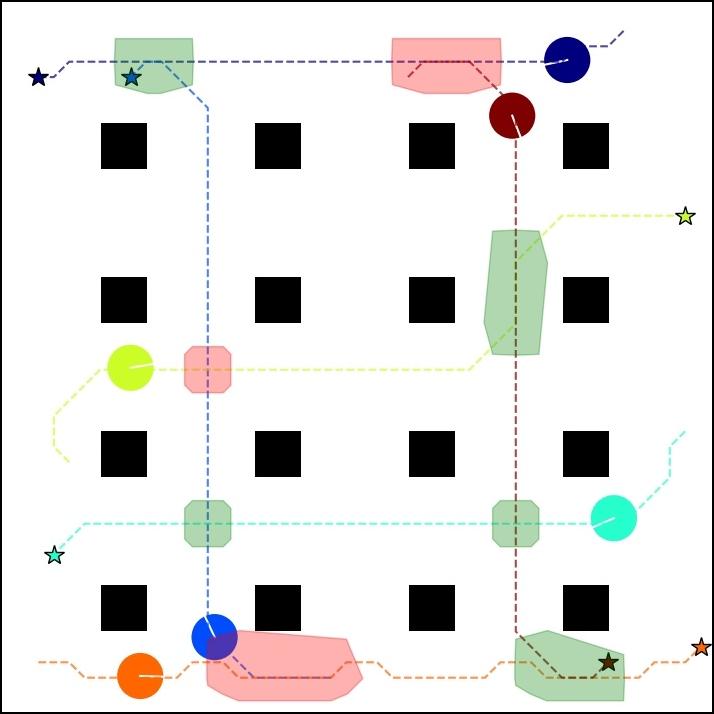}
        \caption{t=4s}
        \label{fig:t4}
    \end{subfigure}
    \hfill
    \begin{subfigure}[b]{0.32\textwidth}
        \includegraphics[width=\textwidth]{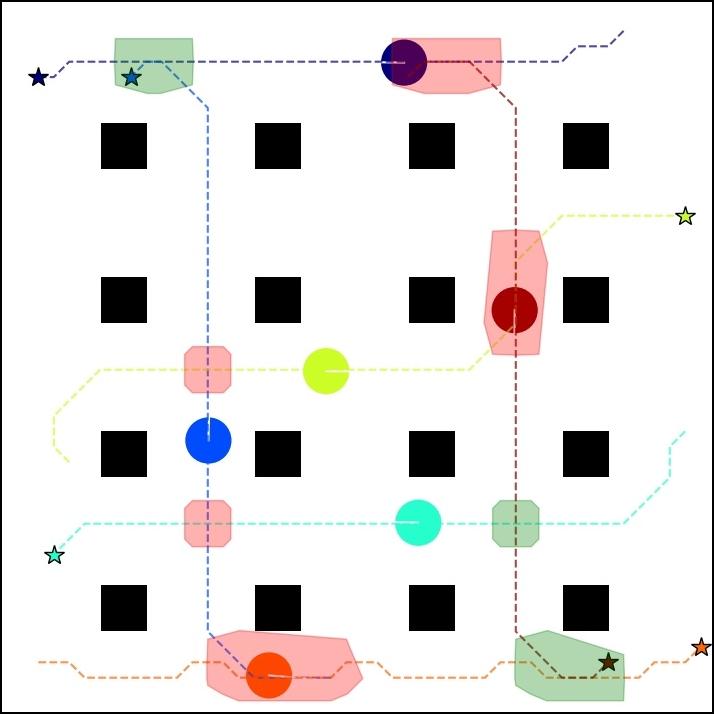}
        \caption{t=8s}
        \label{fig:t8}
    \end{subfigure}

    \vspace{0.5em} 

    \begin{subfigure}[b]{0.32\textwidth}
        \includegraphics[width=\textwidth]{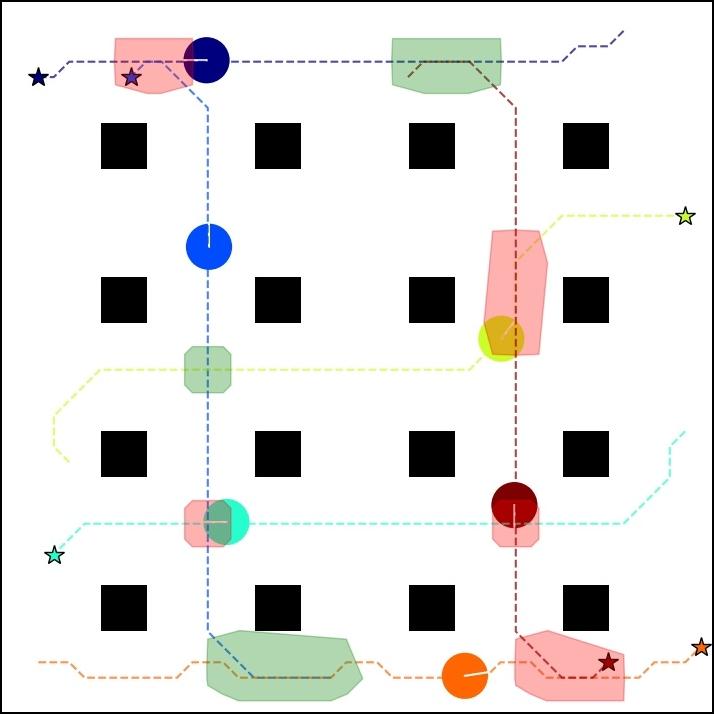}
        \caption{t=12s}
        \label{fig:t12}
    \end{subfigure}
    \hfill
    \begin{subfigure}[b]{0.32\textwidth}
        \includegraphics[width=\textwidth]{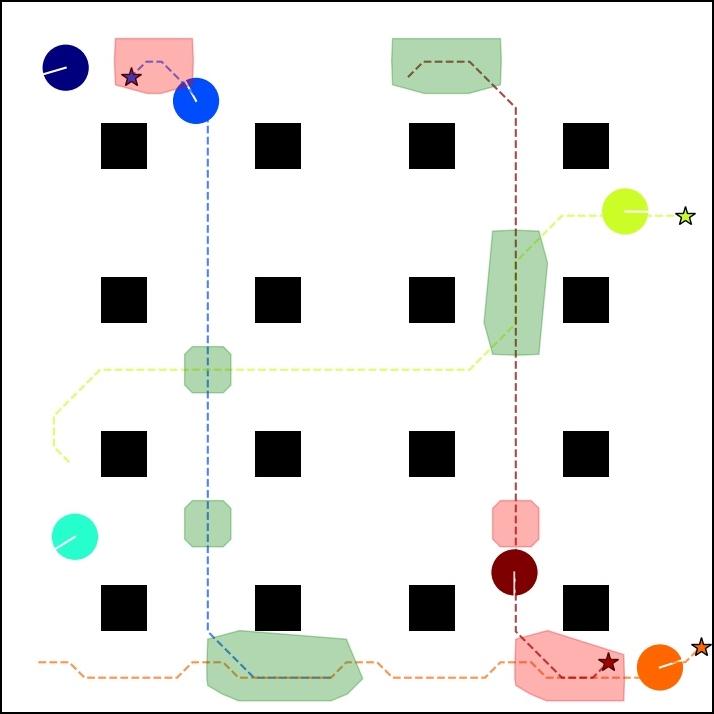}
        \caption{t=16s}
        \label{fig:t16}
    \end{subfigure}
    \hfill
    \begin{subfigure}[b]{0.32\textwidth}
        \includegraphics[width=\textwidth]{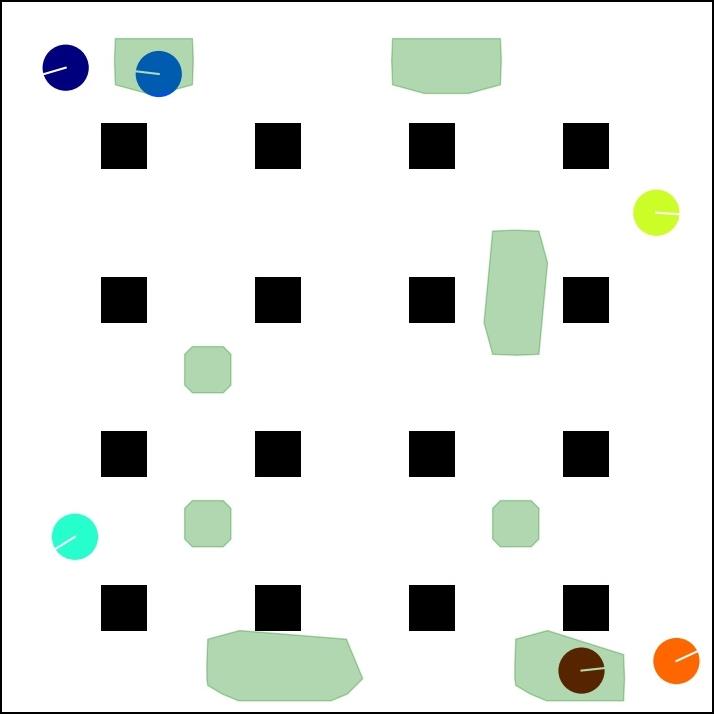}
        \caption{t=20s}
        \label{fig:t20}
    \end{subfigure}
    
    \caption{Snapshots from a simulation run with 6 robots using the hybrid coordination system. There is a 4x4 grid of pillars depicted in black. Robots are depicted as colored circles with a white line representing their orientation. The colored dotted lines depict each robot's global path to their goals, represented by a star of the same color. Intersections are opaque zones which are green when empty and red when occupied.}
    \label{fig:sim_snapshots}
\end{figure*}

Hybrid approaches often combine different algorithms, such as integrating the A* graph search with potential fields for navigation \cite{sang2021hybrid}. \cite{pbm} used a simulation to show that applying a policy framework can effectively regulate robot interactions and resolve conflicts in a hospital setting, while \cite{colregs} adapted maritime collision regulations (COLREGs) to ground robots, where agents independently apply shared traffic rules to resolve conflicts without ambiguity. Moreover, recent work demonstrating a robot navigating intersections by adhering to actual pedestrian traffic lights \cite{trafficlights} highlights the potential for integrating virtual coordination systems, such as the one we are proposing, with physical infrastructure. Our work contributes to this area by proposing a hybrid framework that leverages the strengths of both centralized and decentralized systems, retaining the autonomy of decentralized path planning for individual robots while incorporating a centralized conflict resolution mechanism.

\section{System Overview}

As illustrated in Figure~\ref{fig:sysoverview}, our proposed hybrid system integrates decentralized path planning with centralized conflict management. Each autonomous robot independently computes an optimal path to its goal using an onboard planner and transmits this trajectory to a central server. The server's role is not to plan paths but to act as a conflict mediator; it aggregates the paths from all robots to predict potential collision areas.

Centralized coordination is done using a three-stage process: conflict detection, clustering, and prioritized resolution.\\\textbf{Conflict Detection:} The server continuously performs pairwise checks on the planned paths received from all robots. A conflict is detected if two trajectories intersect and their estimated arrival times are within a predefined threshold. The intersection is then defined by a bounding box that encompasses the overlapping path segments.
\\
\textbf{Conflict Clustering:} When conflicts are detected, the server groups the involved robots into clusters. A cluster contains all robots whose paths are directly or indirectly in conflict. For example, if $R_{1}$ conflicts with $R_{2}$, and $R_{2}$ conflicts with $R_{3}$, all three robots are grouped into a single conflict cluster. 
\\
\textbf{Resolution:} For each conflict zone, resolution occurs at every simulation step. Robots within the zone's larger “stop area” are sorted by proximity to its center to create a priority queue. The system iteratively evaluates this queue, starting with the highest-priority robot. For each subsequent candidate, it checks if its future path conflicts with the paths of all higher-priority robots already cleared to proceed in the current timestep. A “STOP” command is issued if a conflict is detected. Otherwise, the robot can proceed, allowing multiple non-colliding robots to traverse the intersection simultaneously.

\section{Simulation-based Validation}
\subsection{Simulation Setup}
We validated our framework through 1,000 simulations in a Python environment. The world was a $50 \times 50$ unit space containing a $4 \times 4$ grid of static pillar obstacles. Each robot, modeled with a 3-unit diameter, utilized an A* global planner and a Dynamic Window Approach (DWA) local planner. To ensure meaningful navigation, robots in each run were assigned random start and goal positions, with the goal constrained to be at least 75\% of the map's width away from the start.
\begin{figure}
    \centering
    \includegraphics[width=0.99\linewidth]{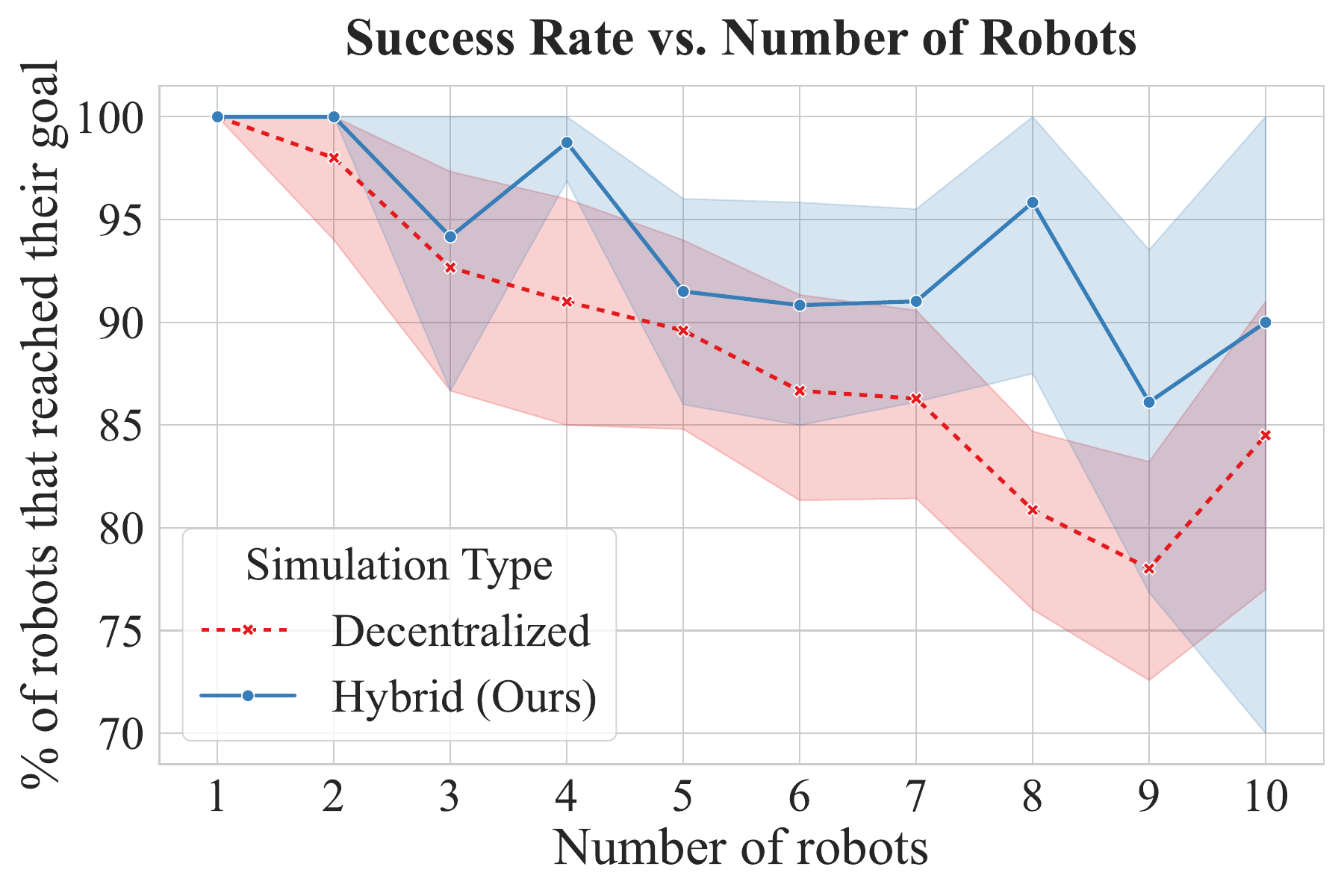}
    \caption{Average success rate (percentage of robots reaching their goal within a 135s timeout) for the Hybrid and Decentralized systems versus the number of robots. Each point is the mean over 500 trials; shaded regions represent the 95\% confidence interval.}
    \label{fig:success-plot}
\end{figure}
We evaluated system performance under three distinct experimental conditions:
\begin{enumerate} 
    \item \textbf{Proposed Hybrid System:} Both the central coordinator and the local DWA collision avoidance were active. Figure~\ref{fig:sim_snapshots} provides a visual representation of a simulation run with six robots under the Hybrid configuration. 
    \item \textbf{Decentralized Baseline:} The central coordinator was disabled. Robots relied exclusively on their local DWA planners, using simulated LIDAR data to avoid collisions with one another. 
\end{enumerate} 

For each configuration, we varied the number of robots from one to ten, running 50 trials per count for a total of 1,000 simulations. Performance was quantified using success rate, total collisions, average speed, and the average number of replans. The success rate is the percentage of robots reaching their goal within a 135-second timeout, a window derived from the maximum potential travel time. The average speed is measured in pixels moved per simulation step. Replans are triggered using a dynamic patience level, varying for each robot from being stuck for 3 to 6 seconds.

\subsection{Results}

\begin{figure}
    \centering
    \includegraphics[width=0.99\linewidth]{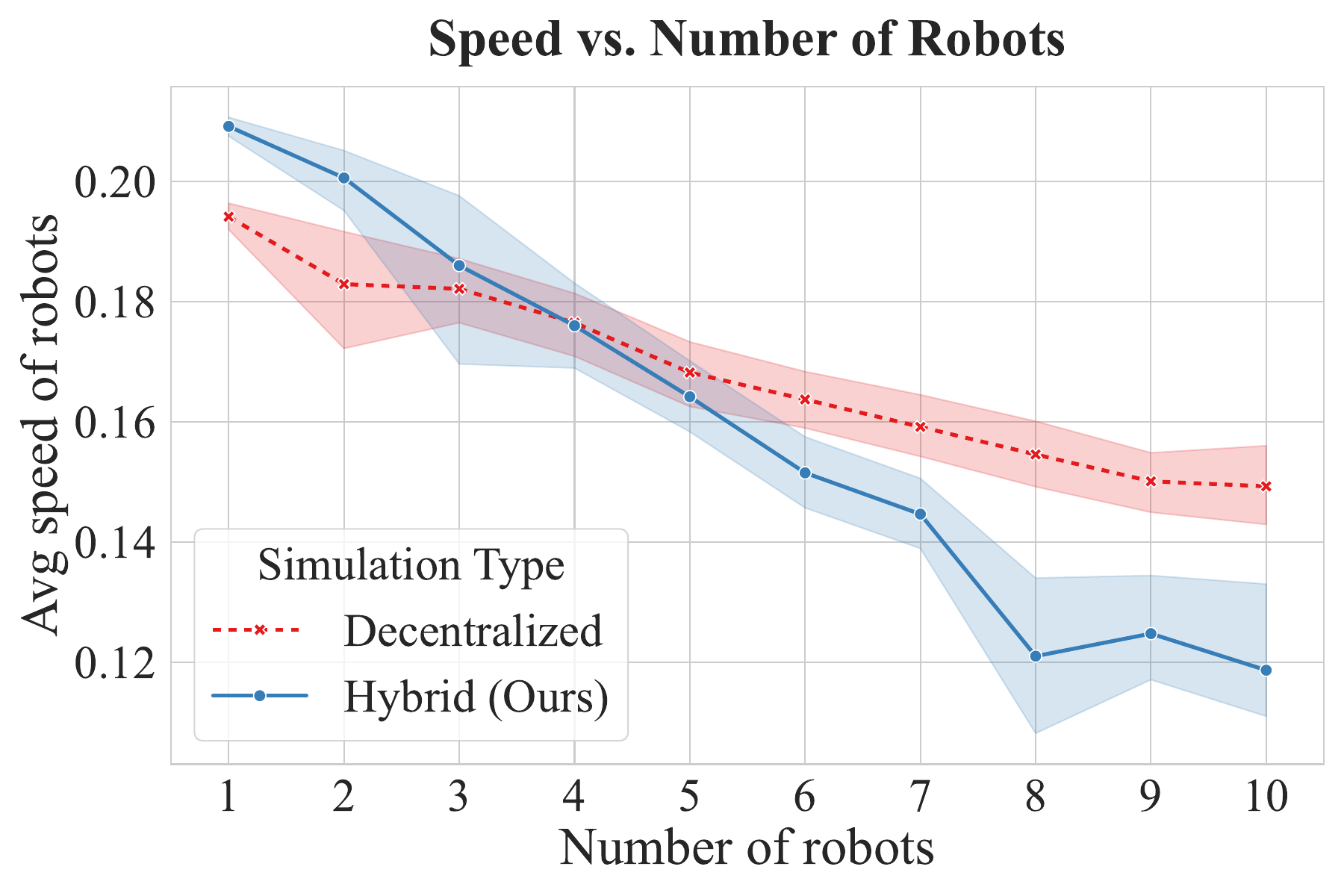}
    \caption{Average speed for Hybrid and Decentralized systems versus the number of robots. Speed is measured in (pixels/step) $\times$ 100, averaged from the start of a run until a robot reaches its goal or times out. Each point is the mean over 500 trials; shaded regions represent the 95\% confidence interval.}
    \label{fig:speed-plot}
\end{figure}

\begin{figure}
    \centering
    \includegraphics[width=0.99\linewidth]{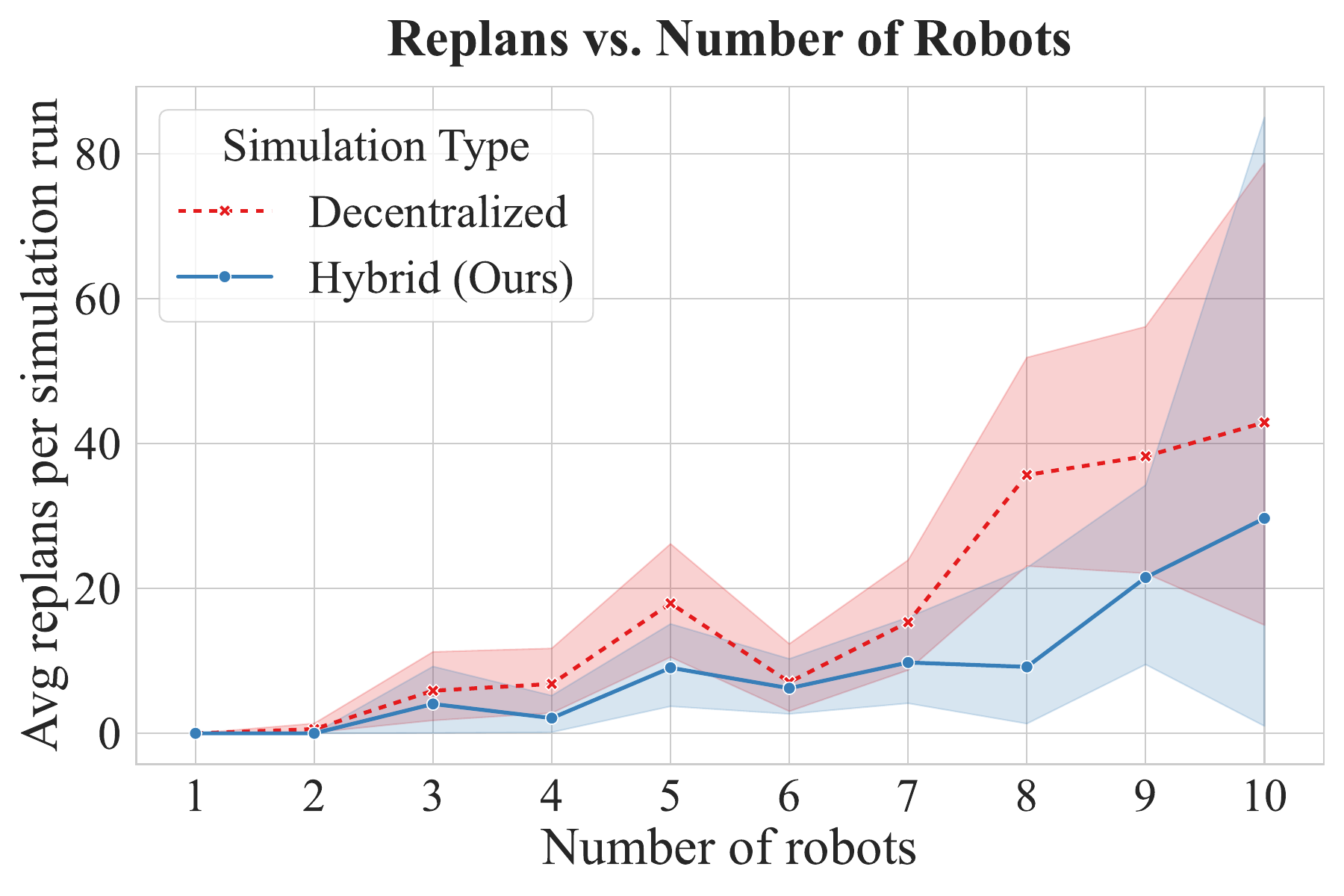}
    \caption{Average total path replans per simulation run for the Hybrid and Decentralized systems versus the number of robots. Each data point is the mean over 500 trials, and the shaded regions indicate the 95\% confidence interval.}
    \label{fig:replan-plot}
\end{figure}

The quantitative results from our 1,000 simulation runs are visualized in the figures below, comparing success rate (Figure~\ref{fig:success-plot}), average speed (Figure~\ref{fig:speed-plot}), and path replans (Figure~\ref{fig:replan-plot}). In addition to these metrics, we monitored collisions between robots and static objects in the environment, and recorded no collisions over 1,000 runs.

Across all multi-robot scenarios (2 to 10 robots), the Hybrid system consistently achieved a higher success rate than the purely Decentralized baseline. The gap in the success rate increased with the number of robots. The Hybrid system's success rate was 96\% compared to the Decentralized system's 81\% with 8 robots, which was the widest difference across the runs. While the success rate for both systems generally decreased with more robots, the Hybrid system's rate remained at or above 90\% until the 9-robot mark. The Decentralized system's rate dropped to as low as 81\% until the 9-robot mark. The Hybrid system was faster with fewer than five robots. The average number of global path replans was lower for the Hybrid system. At eight robots, the Decentralized system required four times as many replans as the Hybrid system.

\subsection{Discussion of Results}

The observed results reveal a clear trade-off between proactive coordination and reactive avoidance. The Hybrid system's success rate is a direct result of its ability to prevent deadlocks. By commanding robots to wait outside a conflict zone, the central coordinator ensures the intersection remains clear for a prioritized robot to pass through. In the purely Decentralized system, robots frequently converge and create gridlock, which is quantified by the high number of replans. These deadlocks cause individual robots to become stuck, ultimately leading them to time out before reaching their goal, which lowers the success rate.
Since decentralized robots are moving until they get stuck, their average speed for successful runs is higher in dense scenarios. The Hybrid system's robots are often stationary, which lowers their average speed but produces a higher rate of task completion. It should be noted that the Hybrid system's effectiveness becomes erratic with 9-10 robots. This uncertainty in performance is statistically represented by the widening confidence interval, which indicates a drop in predictability. The underlying cause for this is that at such high densities, the conflict clustering algorithm begins to merge multiple smaller conflicts into a single, massive, map-spanning intersection. This phenomenon makes the “stop” command less efficient, as robots may be halted far from the actual point of conflict, leading to the high variance in outcomes.

\section{Real World Demonstration}

We validated our framework's applicability through two distinct physical demonstrations, showcasing its flexibility in managing both dynamic and pre-defined conflict zones. We have made available a video of our demonstrations.\footnote{Demonstration video: \url{https://youtu.be/h2lHliLEdd8}}

The first demonstration, shown in Figure~\ref{fig:mainimg}, involved two Unitree GO1 quadruped robots in a $5 \times10m$ laboratory space. Each robot, equipped with a 2D LiDAR, ran ROS 2 and the Nav2 stack on an onboard computer for autonomous mapping and navigation. The robots planned their paths independently and communicated their poses to an external PC running the central coordinator. We created conflict scenarios by assigning start and goal points that resulted in intersecting paths, which allowed the coordinator to dynamically identify and manage the conflict zone.

\begin{figure}
    \centering
    \includegraphics[width=0.99\linewidth]{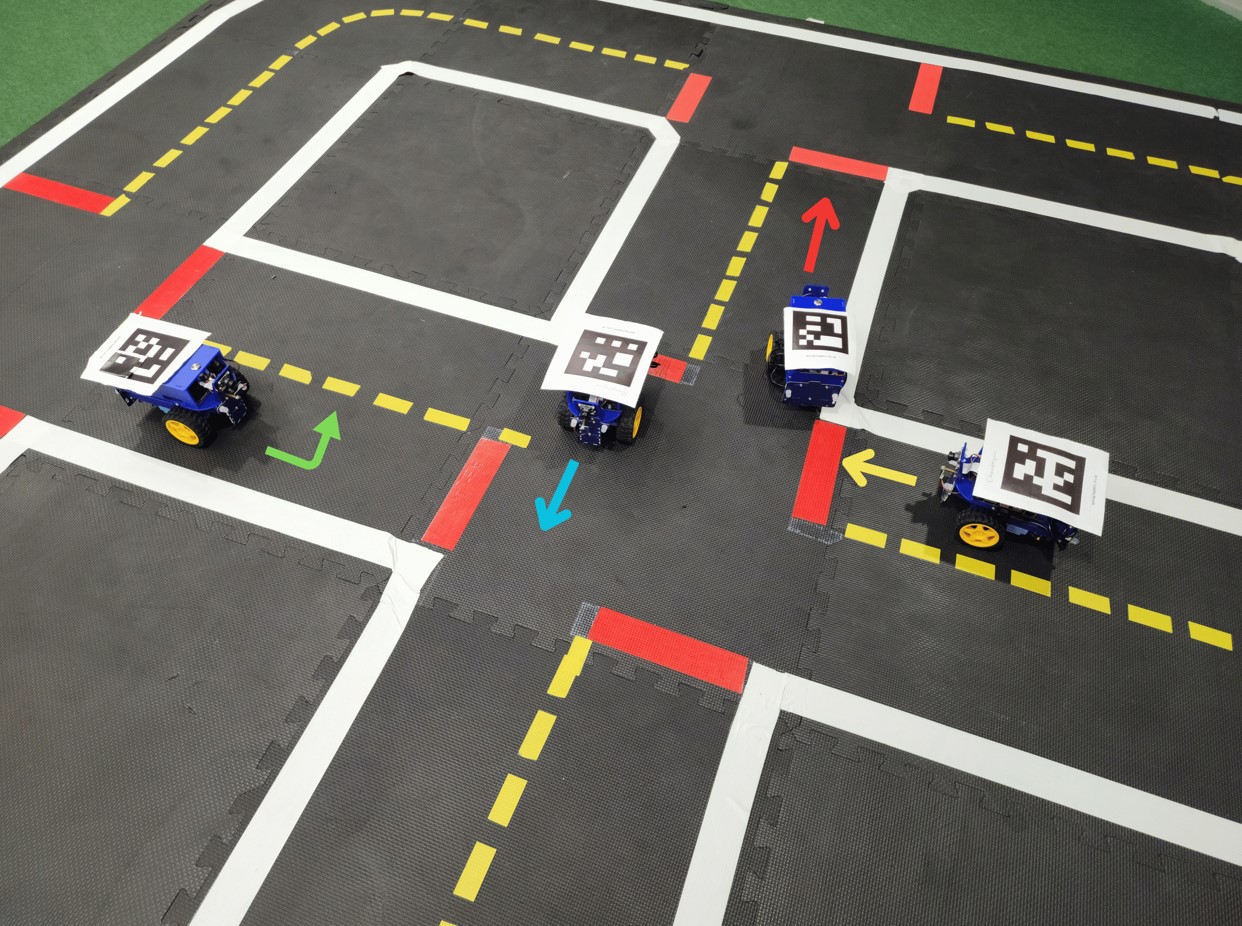}
    \caption{The hybrid system manages a pre-defined intersection with four Duckiebots on a first-come, first-served basis, with arrows indicating turn intentions. The first robot to arrive (red arrow) passes straight, followed in sequence by the blue and yellow robots. The last to arrive (green arrow) is cleared to make its left turn only after the intersection is vacant.}
    \label{fig:duckiedemo}
\end{figure}

The second demonstration, shown in Figure~\ref{fig:duckiedemo}, utilized three Duckiebots doing factory-configured lane-following on a small-scale road network with fixed intersections. A top-down camera system localized the robots using ArUco markers, feeding their positions directly to the central coordinator. In this structured environment, the coordinator's role was to manage access to these pre-defined intersections, granting passage to one robot at a time based on a first-come, first-served policy. This showcased the system's adaptability to scenarios where conflict zones are static and known in advance, and global path information from the robots is not required.

\section{Conclusion}

We presented a hybrid coordination framework that combines decentralized path planning with centralized, conflict resolution, functioning as a virtual traffic light. The system is independent of the planners deployed by independent agents. Simulation results demonstrated that this approach increases goal success rates and reduces path replans by preventing deadlocks when compared to a purely decentralized system, especially in moderately dense scenarios. The framework was also validated in real-world demonstrations with two different robot platforms. Future work will focus on enhancing the central coordinator's conflict mitigation strategies. We plan to enable it to request specific robots to replan their paths, which could improve traffic flow. We also intend to scale up our real-world hardware demonstrations with a larger number of robots. Finally, we aim to extend the simulation framework to 3D environments to evaluate its performance for more applications and a deeper validation. Beyond these directions, our Duckietown experiments also serve as a proof of concept that the same approach could be applied to intersection management for autonomous cars.

\bibliographystyle{named} 
\bibliography{ref.bib}

@article{de2021decentralized,
  title={Decentralized multi-agent pursuit using deep reinforcement learning},
  author={De Souza, Cristino and Newbury, Rhys and Cosgun, Akansel and Castillo, Pedro and Vidolov, Boris and Kuli{\'c}, Dana},
  journal={Robotics and Automation Letters},
  year={2021}
}

@article{effmulti,
  title={Efficient multi-robot path planning in real environments: a centralized coordination system},
  author={Matos, Diogo Miguel and Costa, Pedro and Sobreira, H{\'e}ber and Valente, Antonio and Lima, Jos{\'e}},
  journal={International Journal of Intelligent Robotics and Applications},
  year={2025}
}

@inproceedings{intmulti,
  author    = {Oleiwi, B. K. and Al-Jarrah, R. and Roth, H. and Kazem, B. I.},
  title     = {Integrated motion planing and control for multi objectives optimization and multi robots navigation},
  booktitle = {2nd IFAC Conference on Embedded Systems, Computer Intelligence and Telematics},
  year      = {2015}
}

@article{supermulti,
  author  = {Atinc, G. M. and Stipanovic, D. M. and Voulgaris, P. G.},
  title   = {Supervised coverage control of multi agent systems},
  journal = {Automatica},
  year    = {2014}
}

@article{atzmon,
  author = {Atzmon, Dor and Stern, Roni and Felner, Ariel and Wagner, Glenn and Zhou, Neng-Fa},
  year = {2020},
  title = {Robust Multi-Agent Path Finding and Executing},
  journal = {Journal of Artificial Intelligence Research}
}

@inproceedings{surynek2010optimization,
  title={An optimization variant of multi-robot path planning is intractable},
  author={Surynek, Petr},
  booktitle={Proceedings of the National Conference on Artificial Intelligence},
  year={2010}
}

@article{jingjin2016intractability,
  title={Intractability of optimal multirobot path planning on planar graphs},
  author={Yu, Jingjin},
  journal={IEEE Robotics and Automation Letters},
  year={2016}
}

@book{gunter2014history,
  title={The history of automated guided vehicle systems},
  author={G{\"u}nter, Ullrich},
  year={2014}
}

@article{siefke2020robotic,
  title={Robotic systems of systems based on a decentralized service-oriented architecture},
  author={Siefke, Lars and Sommer, Volker and Wudka, Benedikt and Thomas, Christian},
  journal={Robotics},
  year={2020}
}

@inproceedings{flocchini2000distributed,
  title={Distributed coordination of a set of autonomous mobile robots},
  author={Flocchini, Paola and Prencipe, Giuseppe and Santoro, Nicola and Widmayer, Peter},
  booktitle={Intelligent Vehicles Symposium, Proceedings},
  year={2000}
}

@inproceedings{jouandeau2012decentralized,
  title={Decentralized waypoint-based multi-robot coordination},
  author={Jouandeau, Nicolas and Yan, Zheng},
  booktitle={International Conference on Cyber Technology in Automation, Control and Intelligent Systems, Bangkok, Thailand},
  year={2012}
}

@article{iocchi2003distributed,
  title={Distributed coordination in heterogeneous multi-robot systems},
  author={Iocchi, Luca and Nardi, Daniele and Piaggio, Michele and Sgorbissa, Antonio},
  journal={Autonomous Robots},
  year={2003}
}

@inproceedings{caloud1990indoor,
  title={Indoor automation with many mobile robots},
  author={Caloud, Philippe and Wonyun Choi, J-C and Latombe, C and Pape, Yim, M},
  booktitle={International Workshop on Intelligent Robots and Systems, Towards a New Frontier of Applications},
  year={1990}
}

@inproceedings{berndt2021centralized,
  title={Centralized Robotic Fleet Coordination and Control},
  author={Berndt, Michael and Krummacker, Dennis and Fischer, Christian and Schotten, Hans D},
  booktitle={Mobile Communication-technologies and applications},
  year={2021}
}

@article{sang2021hybrid,
  title={The hybrid path planning algorithm based on improved {A*} and artificial potential field for unmanned surface vehicle formations},
  author={Sang, Hongke and You, Yingtang and Sun, Xiujun and Zhou, Yang and Liu, Fang},
  journal={Ocean Engineering},
  year={2021}
}

@misc{kulathunga2021reinforcement,
  title={A reinforcement learning based path planning approach in 3D environment},
  author={Kulathunga, Gayanga},
  year={2021}
}

@article{junyan2020voronoi,
  title={Voronoi-based multi-robot autonomous exploration in unknown environments via deep reinforcement learning},
  author={He, Junyan and Niu, Hanlin and Carrasco, Joaquin and Lennox, Barry and Arvin, Farshad},
  journal={Transactions on Vehicular Technology},
  year={2020}
}

@article{zagradjanin2021cloud,
  title={Cloud-based multi-robot path planning in complex and crowded environment using fuzzy logic and online learning},
  author={Zagradjanin, Nenad and Rodic, Aleksandar and Pamucar, Dragan and Pavkovic, Branimir},
  journal={Information Technology and Control},
  year={2021}
}

@INPROCEEDINGS{pbm,
  author={Batool, Amna and Loke, Seng W. and Fernando, Niroshinie and Kua, Jonathan},
  booktitle={2024 IEEE Smart World Congress}, 
  title={Policy-based Management of Human-Device and Device-Device Interactions in {IoT} Collectives: A Simulation-based Study}, 
  year={2024}
}

@INPROCEEDINGS{colregs,
  author={Jha, Eshant and Somayajula, Abhilash and Gideon, Don and Raveendran, Sayooj P and Sebastian, Bijo},

  booktitle={2024 IEEE International Conference on Artificial Intelligence in Engineering and Technology}, 

  title={{COLREGs} Inspired Decentralised Path Planning for Multi-Agent System}, 

  year={2024}
}

@INPROCEEDINGS{trafficlights,
  author={Gupta, Sagar and Cosgun, Akansel},
  booktitle={2024 IEEE/RSJ International Conference on Intelligent Robots and Systems (IROS)}, 
  title={Audio-Visual Traffic Light State Detection for Urban Robots}, 
  year={2024},
  volume={},
  number={},
  pages={12509-12514},
  doi={10.1109/IROS58592.2024.10802855}}

\end{document}